\documentclass[runningheads]{llncs}
\usepackage[T1]{fontenc}
\usepackage{graphicx}
\usepackage[table, dvipsnames]{xcolor}
\usepackage{amsmath}
\usepackage[hidelinks]{hyperref}
\usepackage[sort&compress,capitalize,noabbrev,nameinlink]{cleveref}
\usepackage[acronym]{glossaries}
\usepackage{graphicx}
\usepackage{subcaption}

\creflabelformat{equation}{#2#1#3}

\newacronym{snn}{SNN}{Spiking Neural Network}
\newacronym{ann}{ANN}{Artifical Neural Network}
\newacronym{dl}{DL}{Deep Learning}
\newacronym{lif}{LIF}{Leaky Integrate-and-Fire}
\newacronym{if}{IF}{Integrate-and-Fire}
\newacronym{stdp}{STDP}{Spike Timing Dependent Plasticity}
\newacronym{dvs}{DVS}{Dynamic Vision Sensor}
\newacronym{bptt}{BPTT}{Back-Propagation Through Time}
\newacronym{sgd}{SGD}{Stochastic Gradient Descent}
\newacronym{rvt}{RVT}{Recurrent Vision Transformer}
\newacronym{tbptt}{TBPTT}{Truncated Backpropagation Through Time}

\begin{document}
\title{ShapeAug++: More Realistic Shape Augmentation for Event Data}
%
%
\author{Katharina Bendig \inst{1} \and
René Schuster \inst{1,2}\and
Didier Stricker \inst{1,2}}
\authorrunning{K. Bendig et al.}
%
\institute{RPTU - University Kaiserslautern-Landau, Germany \and
DFKI - German Research Institute for Artifical Intelligence, Germany \\
\email{firstname.lastname@dfki.de}}
\maketitle              
\begin{abstract}

The novel \Glspl*{dvs} gained a great amount of attention recently as they are superior compared to RGB cameras in terms of latency, dynamic range and energy consumption. This is particularly of interest for autonomous applications since event cameras are able to alleviate motion blur and allow for night vision. One challenge in real-world autonomous settings is occlusion where foreground objects hinder the view on traffic participants in the background. The ShapeAug method addresses this problem by using simulated events resulting from objects moving on linear paths for event data augmentation. However, the shapes and movements lack complexity, making the simulation fail to resemble the behavior of objects in the real world. Therefore in this paper, we propose ShapeAug++, an extended version of ShapeAug which involves randomly generated polygons as well as curved movements. We show the superiority of our method on multiple \gls{dvs} classification datasets, improving the top-1 accuracy by up to 3.7\% compared to ShapeAug.
\keywords{Event Camera Data \and Augmentation \and Classification.}

\end{abstract}

\section{\uppercase{Introduction}}
\label{sec:introduction}
\glsresetall

\begin{figure}[htbp]
    \centering
    \begin{subfigure}{0.32\textwidth}
        \includegraphics[width=\textwidth]{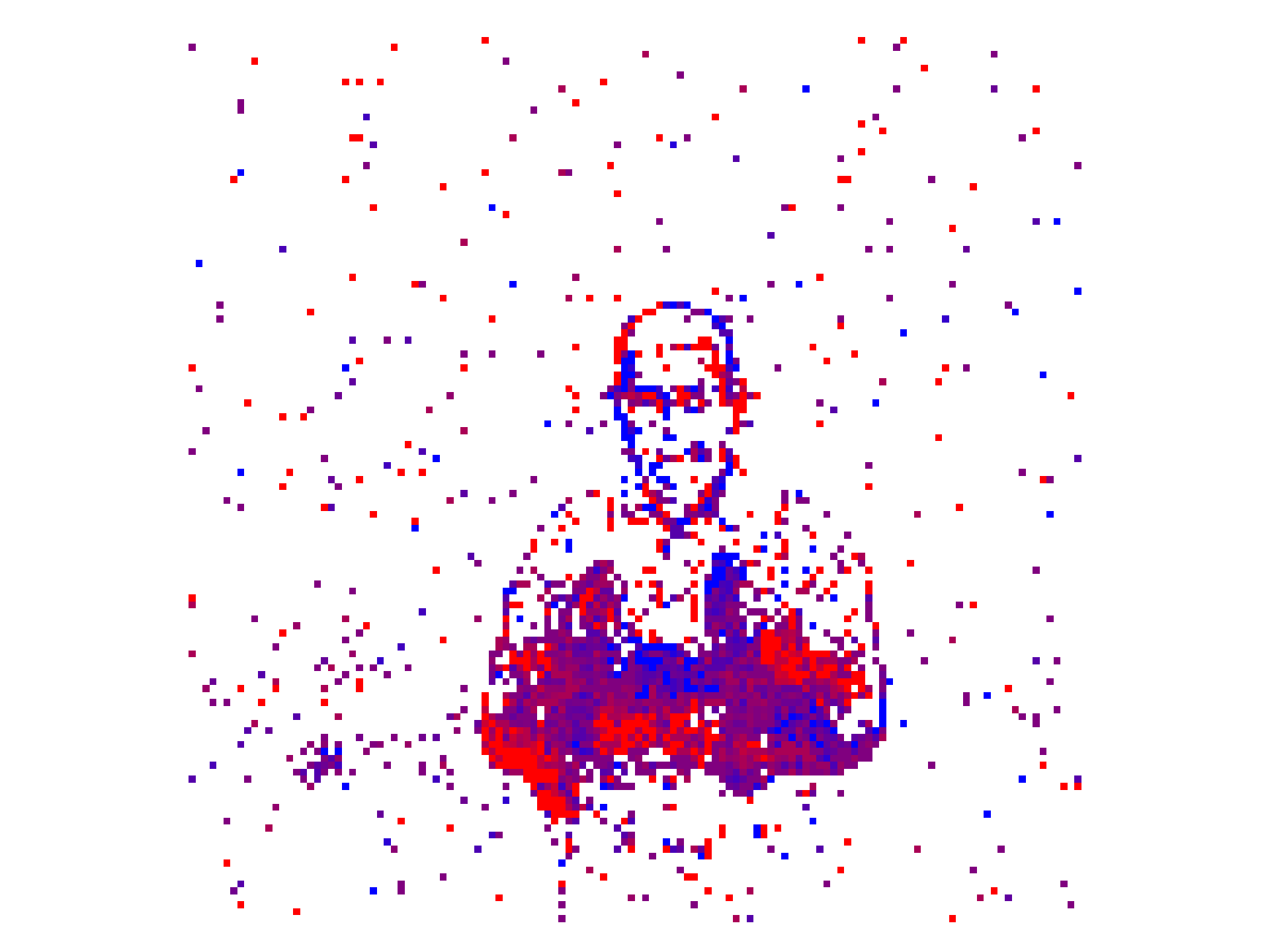}
        \caption{Original}
    \end{subfigure}
    \hfill
    \begin{subfigure}{0.32\textwidth}
        \includegraphics[width=\textwidth]{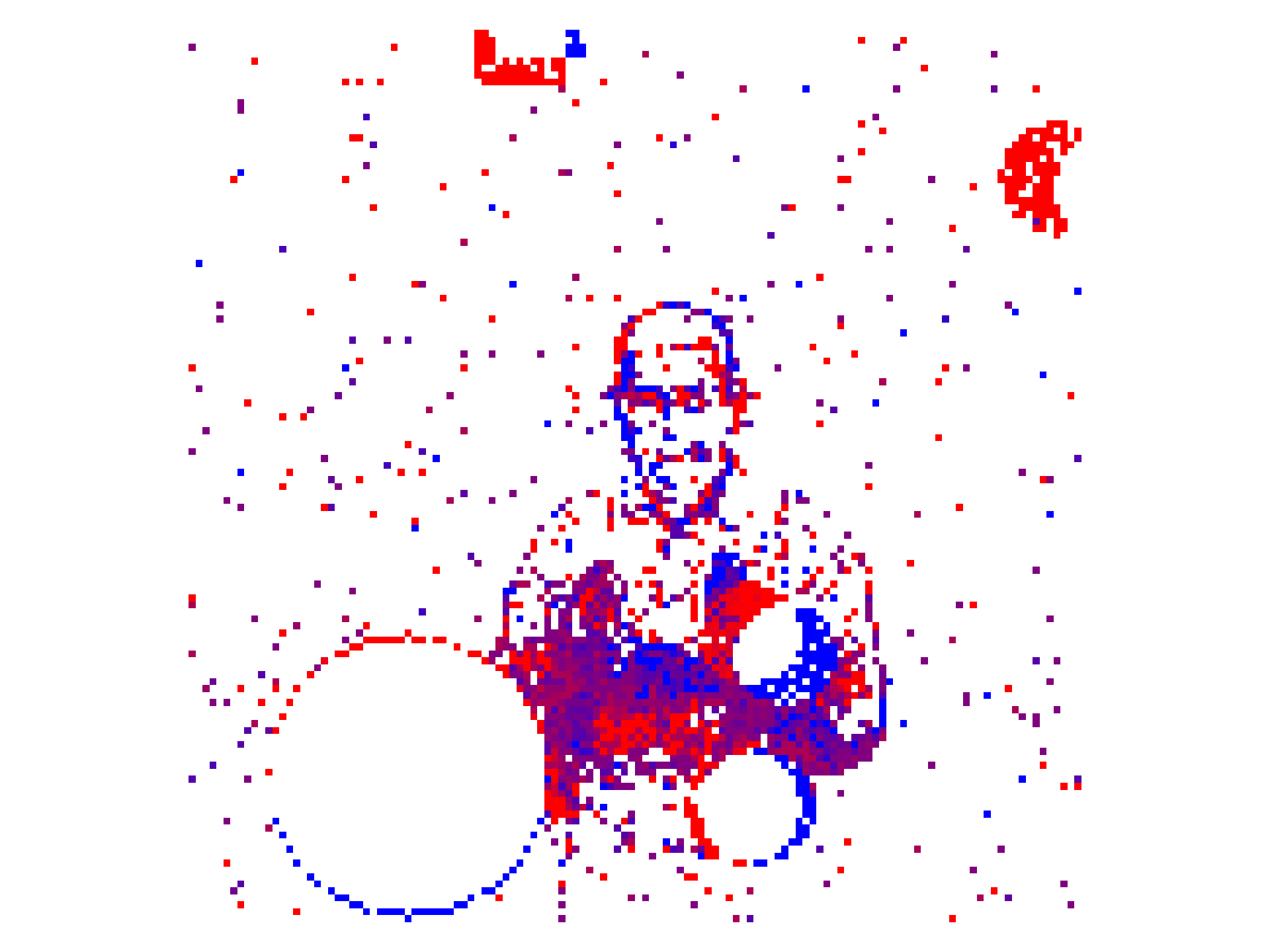}
        \caption{ShapeAug}
    \end{subfigure}
    \hfill
    \begin{subfigure}{0.32\textwidth}
        \includegraphics[width=\textwidth]{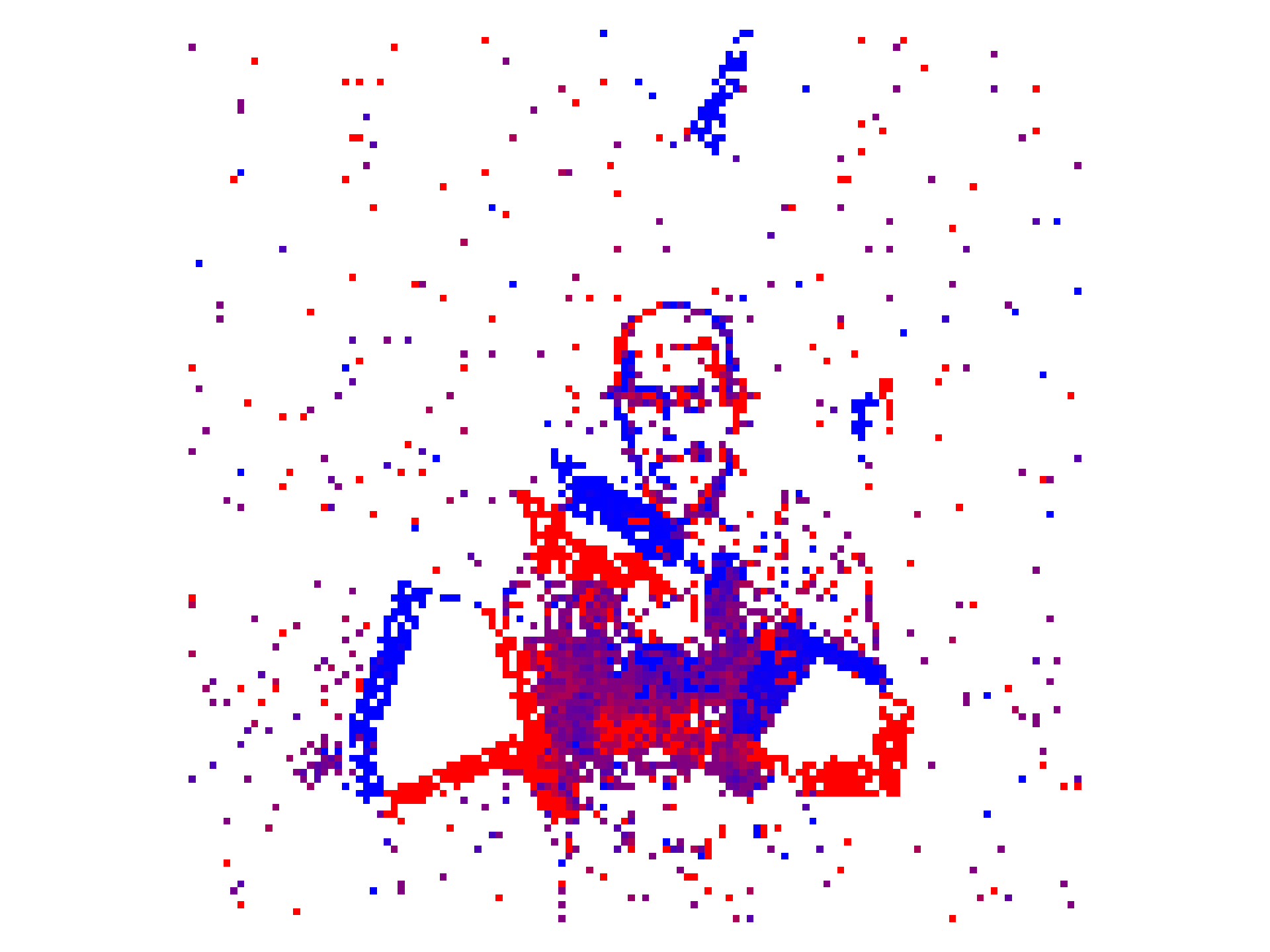}
        \caption{ShapeAug++}
    \end{subfigure}
    \caption{Visualization of the ShapeAug and ShapeAug++ augmentation methods on \textit{DVS-Gesture} \cite{dvsgesture}.}
    \label{fig:aug_comparison}
\end{figure}

The field of autonomous driving advances rapidly and has the potential to vastly change everyday life for a great amount of people. Therefore, it is imperative that the robustness and safety of this technology has to be a priority. One promising progression is the development of \Glspl*{dvs}, also known as event cameras. These vision sensors, unlike unconventional RGB cameras, register brightness changes asynchronously instead of absolute intensity values at a pre-defined frame rate. This enables them to record visual input with an exceedingly low latency, in the range of milliseconds. \Glspl*{dvs} therefore allow a crucially fast detection of other traffic participants, which are able to move multiple meters in mere seconds. In addition, event cameras have a high dynamic range (in the range of 140 dB) and can thus record motion information even at night time and during poor lighting conditions. Due to their asynchronous pixels, \Glspl*{dvs} moreover have a low energy consumption, allowing for their utilization in any mobile application and leading to a comparably low carbon footprint. 

However, given that this is a relatively recent technology, the amount of data available for \Gls*{dl} approaches is quite limited. In comparison to RGB datasets like ImageNet \cite{ImageNet}, which contains 14 million images, event datasets such as N-CARS \cite{HATS} have only a few thousand labeled images. Consequently, data augmentation becomes crucial to prevent overfitting and increase the robustness of neural networks. Even with larger datasets like the Gen1 Automotive Event Dataset \cite{gen1}, other works have shown that simple geometric augmentation methods can significantly boost the performance by up to 25\% \cite{RVT}.

Especially in the context of autonomous driving systems, occlusion is another great challenge for \Gls*{dl} detection methods. Occlusion describes the (partial) covering of background objects by other foreground items. In order to avoid dangerous accidents, detection methods must be capable to detect these occluded objects nevertheless. Since even the labeling of occluded items is challenging, data augmentation methods become crucial to ensure the robustness and sufficient detection by neural networks. Many augmentation methods for handling occlusion like \cite{eventdrop,dropout} typically remove data over time or in specific areas. However, this can only simulate an occluding object moving synchronously with the camera, which is not representative of real-world scenarios. 
Automotive settings are inherently dynamic, meaning most objects within the scene are in motion independent of the camera's ego-motion, which cannot be accurately represented by simply dropping events at a fixed location.

For this reason, we have introduced a method for the more realistic simulation of occluding objects which move in the foreground in our work ShapeAug \cite{shapeaug}. To do so, random objects in the form of squares and circles are generated, which move along linear paths into a random direction. Moreover, the resulting events from this movement are simulated and applied into the foreground of the scene. Our experiments show, that ShapeAug is able to increase the performance for classification tasks as well as the robustness against various challenging validation data. 

However, perfect squares and linear movements are rarely encountered in real-world scenarios. To further improve performance and realism in simulated occlusions, we introduce \textbf{ShapeAug++}. This advanced method significantly increases the complexity of simulated objects and their movements. \textbf{ShapeAug++} can generate various polygons by using the convex hull of randomly distributed points. It also incorporates arbitrary Bézier curves for movement and includes object rotations around their own axes. We show that our improved augmentation technique is able to outperform ShapeAug on the most common event datasets for classification.

\section{\uppercase{Related Work}}
\label{sec:relatedwork}

\subsection{Occlusion-aware RGB Image Augmentation}

One common approach to augmenting RGB images, which can be viewed as occlusion simulation, involves removing (zeroing out) specific regions of the image. Hide-and-Seek \cite{hideandseek} is one such method, which divides the image into a fixed number of patches and assigns each patch a probability of being removed. Alternatively, the Cutout method \cite{cutout} abandons the rigid grid structure by randomly selecting a fixed number of center points in the image and removing squares of a predefined side length around these points. Building on these methods, the work by \cite{occ_img_cls} incorporates a gradient-based saliency approach along with Batch Augmentation \cite{BatchAug}. However, these methods do not effectively replicate real-world event occlusion, where occlusion in consecutive frames needs temporal correlation, and moving foreground objects would generate events themselves.

\subsection{Event Data Augmentation}

Event data augmentation techniques often stem from methods used for RGB images. For example, \cite{NeuroDataAug} applies various geometric augmentations -- such as horizontal flipping, shifting, rotation as well as Cutout \cite{cutout} and CutMix \cite{cutmix}. As in our previous work ShapeAug \cite{shapeaug}, we apply geometric augmentation during all our experiments. 

CutMix combines two samples and their labels through linear interpolation. EventMix \cite{eventmix} applies this idea to event data but falls short in realistically modeling occlusion, as it does not account for situations where a foreground object fully covers a background object.

Inspired by Dropout \cite{dropout}, the work of \cite{eventdrop} introduces EventDrop, which drops events randomly based on time and area. EventRPG \cite{eventrpg} extends this idea by computing saliency maps for Spiking Neural Networks allowing for a relevance based application of EventDrop and EventMix. These approaches, however, are unable to adequately simulate occlusion in dynamic real-world scenes because moving objects would still generate events unless perfectly synchronized with the camera. 

Our previous method ShapeAug \cite{shapeaug} overcomes this limitation by simulating both the occlusion caused by foreground objects and the events resulting from their movement. However, the simplistic nature of the objects (squares and circles) and their linear movements does not capture the complexity of natural shapes and motion. In this paper, we build upon ShapeAug by introducing more complex shapes based on polygons as well as curved movements and rotations.

\section{Method}

\subsection{Event Data Handling}

Following the example of ShapeAug, we construct event histograms $E$ based on the asynchronous events $e_i = (x_i, y_i, t_i, p_i)$ consisting of the pixel position $(x_i, y_i)$ and time $t_i$ of an event as well as its polarity $p_i$. The event histogram $E$ has the shape $(T, 2, H, W)$ with $(H,W)$ as the height and width of the event sensor and $T$ as the number of timesteps one event sample is split into. Mathematically, $E$ can be constructed in the following manner:
\begin{align}
    \mathcal{E}(\tau, p, x,y) &= \sum_{e_i \in \mathcal{E}} \delta(\tau - \tau_i) \delta(p - p_i) \delta(x - x_i) \delta(y - y_i), \\
    \tau_i &=  \left \lfloor{\frac{t_i - t_a}{t_b -t_a}\cdot T}\right \rfloor,
\end{align}
with $\delta(\cdot)$ as the Kronecker delta function.

In our classification experiments, the timesteps are fed sequentially into the network, as we are utilizing a \gls{snn}.

While we use event histograms due to their popularity for densifying events and simplifying neural network handling, our method can theoretically work with any event representation.

\subsection{Object Generation}

\begin{figure} [t]
    \centering
    \begin{subfigure}{0.32\textwidth}
        \includegraphics[width=\textwidth]{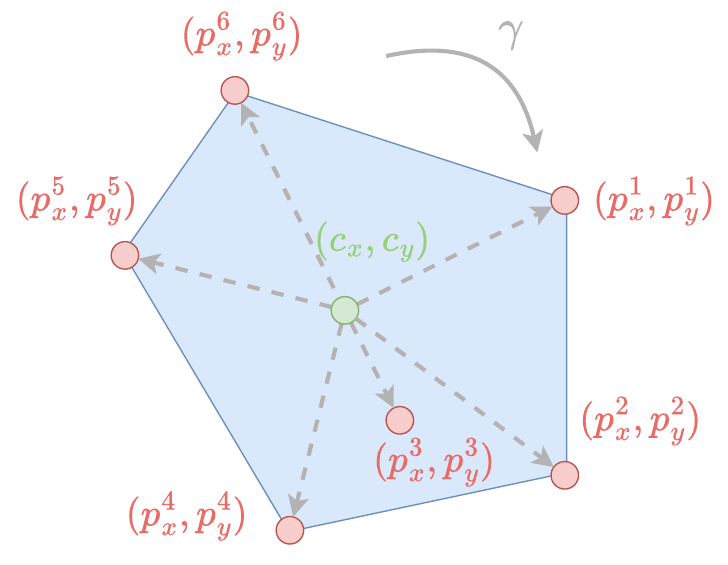}
        \caption{Shape Generation}
    \end{subfigure}
    \hfill
    \begin{subfigure}{0.32\textwidth}
        \includegraphics[width=\textwidth]{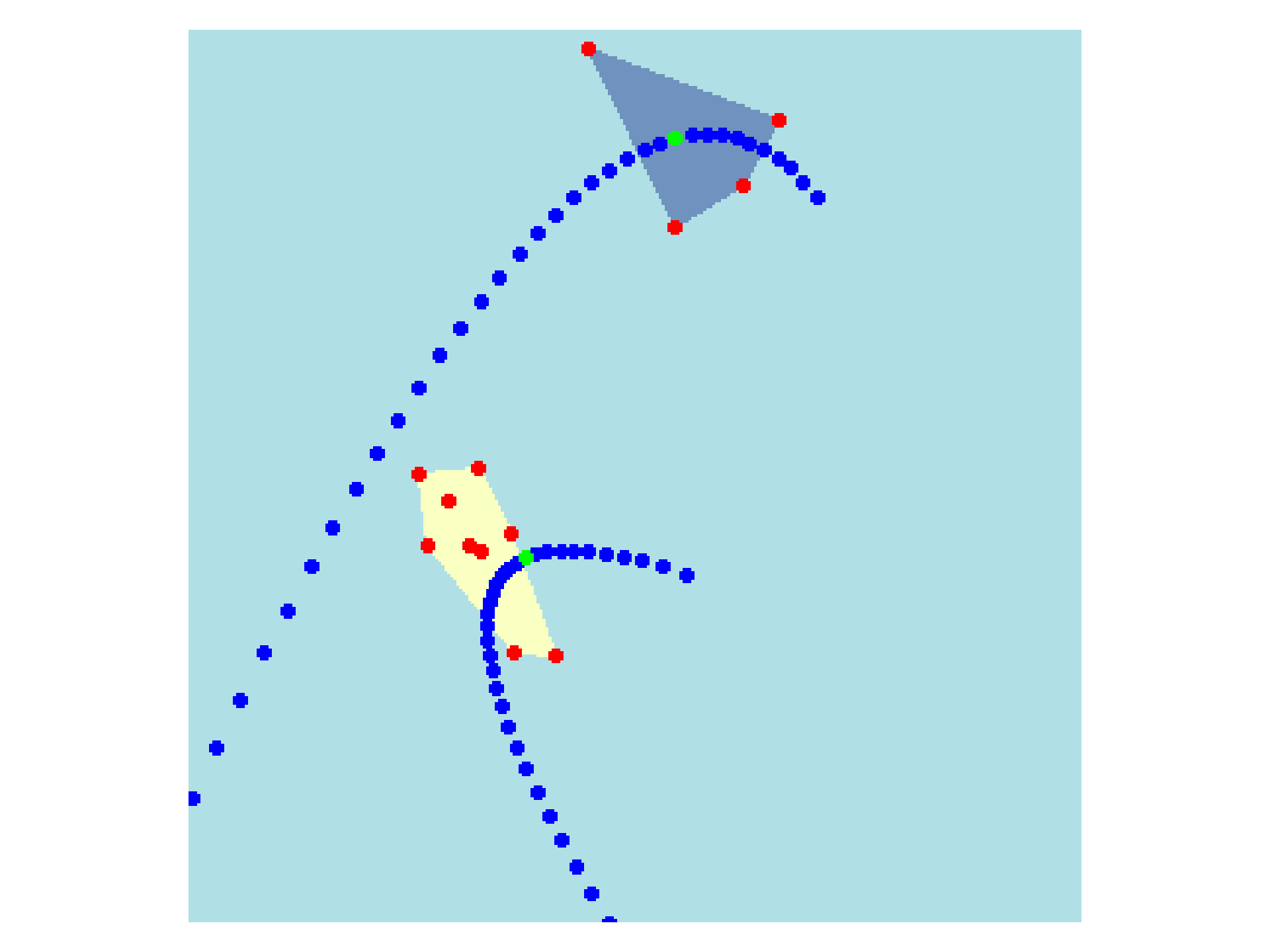}
        \caption{Example Frame}
    \end{subfigure}
    \hfill
    \begin{subfigure}{0.32\textwidth}
        \includegraphics[width=\textwidth]{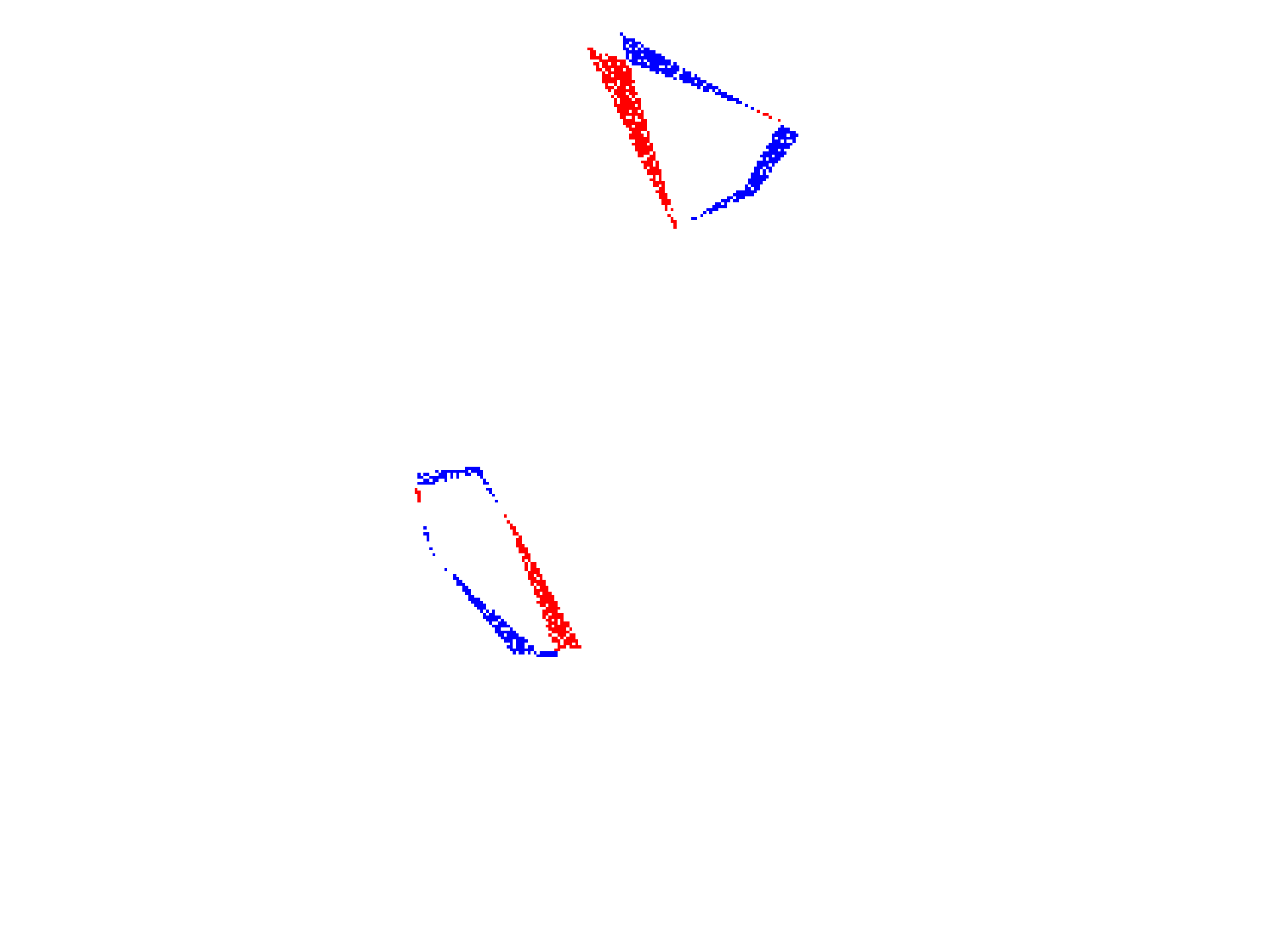}
        \caption{Resulting Events}
    \end{subfigure}
    \caption{Visualization of (a) the shape generation process, (b) an example generated frames for ShapeAug++ and (b) the resulting events. The path along the Bézier curves is illustrated with blue points, the center point is highlighted in green, and the offset points are marked in red. These points are displayed here for visualization purposes only.}
    \label{fig:process}
\end{figure}

To ensure a wide variety of complex shapes, we employ a method to generate objects as random polygons, which is visualized in \cref{fig:process}. The number of shapes $n_s$ is a random integer sampled from a uniform distribution between 1 and the maximum number of shapes $N_s$. Then for each shape a random center point $(c_x,c_y)$ within the frame with dimensions $W,H$ is selected. The coordinates of the center point are drawn from uniform distributions: 
\begin{equation}
    (c_x,c_y)=(\mathcal{U}(0,W - 1),\mathcal{U}(0,H - 1)).
\end{equation}
For each shape, we generate a random number of vertices $N_p$, which is sampled uniformly between 6 and 10. The position of each vertex is defined relative to the center point and is calculated as:
\begin{equation}
    (p_x,p_y)=(c_x + \mathcal{U}(- s, s),c_y + \mathcal{U}(- s,s)),
\end{equation}
with $s = \mathcal{U}(s_{min}, s_{max})$ as the size for the shape.

The polygons are constructed by computing the convex hull of the set of vertices so that the shapes are still big enough to cause occlusion. Moreover each polygon as well as the background are assigned a random color, contributing to the diversity of the generated shapes and introducing more randomness to the polarity of events.

\subsection{Object Movement}

To simulate more complex and natural movements, we generate the paths of the objects using random quadratic Bézier curves. Each curve is defined by three points: The starting point $P_0 = (c_x, c_y)$, the control point \\ 
$P_1 = (\mathcal{U}(-\frac{W}{2},\frac{W}{2}),\mathcal{U}(-\frac{H}{2},\frac{H}{2}))$, and the end point $P_2$, which is a random point outside the frame by a margin of $s$ to ensure the shape moves out of the frame.
To compute $n$ points on the quadratic Bézier curve, we can utilize the following equations:
\begin{align}
t_i &= \frac{i}{n-1}, \quad \text{for each } i \in [0, n-1], \\
\mathbf{B}(t_i) &= (1-t_i)^2 \mathbf{P}_0 + 2(1-t_i)t_i \mathbf{P}_1 + t_i^2 \mathbf{P}_2.
\end{align}
For each frame, the center point of the object is set to the next point $\mathbf{B}(t_i)$ on the curve. Additionally, for each shape, we randomly choose an angle $\gamma = \mathcal{U}(-10,10)^{\circ}$ and rotate it by this angle at each step.

These complex movements result in significant variety, especially in the events which are formed on different sides of the objects depending on their direction of movement. Moreover, due to the non-uniform distribution of points along a Bézier curve, the objects can also accelerate or decelerate along their paths.

\subsection{Frame-based Event Simulation}

To keep the computational overhead of our augmentation approach low, we use the same method as ShapeAug to simulate events based on frames. The first step involves generating frames with various moving objects, which were detailed in the previous sections. We then compute the difference between consecutive frames, as \glspl{dvs} register changes in lighting rather than absolute intensity values. Positive and negative differences correspond to positive and negative events, respectively.

To enhance realism, we simulate noise by randomly varying the number of events per pixel and setting some to zero. We also clip all values to 0.9 times the maximum of all non-zero values times a small random value. Additionally, pixels behind generated foreground objects are masked out to simulate occlusion.

\section{\uppercase{Experiments and Results}}
\label{sec:experiments}

\subsection{Datasets}

For validation, we utilize the same datasets and settings as the original ShapeAug, which include two image-based event datasets and two original event camera datasets for classification. The converted datasets were created by recording RGB images from existing RGB datasets using an event camera. One such dataset is \textit{DVS-CIFAR10} \cite{cifar10dvs}, the event-based version of CIFAR-10 \cite{cifar10}, containing 10,000 event streams at a resolution of $128px \times 128px$. Similarly, \textit{N-Caltech101} \cite{ncaltech101}, derived from Caltech101 \cite{caltech101}, consists of 8,709 images of varying sizes. Among the real event datasets,\textit{ N-CARS} \cite{HATS} is designed for vehicle classification, containing 15,422 training samples and 8,607 test samples at a resolution of $120px \times 100px$ pixels. Another real-world dataset, \textit{DVS-Gesture} \cite{dvsgesture}, focuses on gesture recognition and includes samples of 11 hand gestures performed by 29 subjects, resulting in 1,342 samples at a resolution of $128px \times 128px$. Following the example of \cite{eventmix}, we resize all event streams to $80px \times 80px$ pixels using bi-linear interpolation and divide them into 10 timesteps. For datasets without a predefined training-validation split, we utilize the same split as \cite{eventmix}.
We compare our method (Shape++) against ShapeAug \cite{shapeaug} (Shape) and the two other leading state-of-the-art event augmentation techniques: EventDrop \cite{eventdrop} (Drop), which randomly masks out events in specific areas or time intervals, and EventMix \cite{eventmix} (Mix), which combines multiple samples into a single augmented sample.

\subsection{Implementation}

Following the approach in ShapeAug \cite{shapeaug}, we use the same training settings as \cite{eventmix} for all our classification experiments on DVS-CIFAR10, N-Caltech101, N-CARS, and DVS-Gesture. Specifically, we employ a preactivated spiking ResNet34 \cite{PreActResnet} with PLIF neurons \cite{PLIF}, and the AdamW optimizer \cite{AdamW} with a learning rate of $1.56 \times 10^{-4}$ and a weight decay of $1\times 10^{-4}$. The model is trained using a batch size of 32 for 200 epochs with a cosine decay applied to the learning rate.
For our baseline, we use geometric augmentation, which includes cropping to $80px \times 80px$ pixels after padding with 7 pixels, random horizontal flipping, and random rotation up to $15^{\circ}$.

\begin{table*}
\caption{Comparison of classification results using different max shape sizes $s_{max}$ on all four classification datasets. We report the top-1 accuracy as well as the top-5 accuracy in parantheses, except for datasets with less than 5 classes. The best and second best results are shown in \textbf{bold} and \underline{underlined} respectively. }
\label{tab:classification}
\begin{center}
\begin{tabular}{cccccc}
 \hline 
Method  & $s_{max}$ [px] & DVS-CIFAR10 & N-Caltech101 & N-CARS & DVS-Gesture \\
 \hline 
 Geo  & - & 73.8 (95.5) & 62.2 (81.5) & 97.1 &89.8 (99.6)\\
  \hline
 Geo + ShapeAug  & 10 & 74.3 (95.1) & 68.0 (83.9)& \underline{97.3} & 90.9 (99.6)\\
 Geo + ShapeAug  & 30 & 73.9 (94.7)& 68.7 (86.9) & 96.9 &\underline{91.7} (100)\\
 Geo + ShapeAug & 50 &\textbf{75.7} (96.7)& 68.2 (85.2)& 96.9 &90.5 (99.2)\\
  \hline
   Geo + ShapeAug++  & 10 & 74.1 (95.8) & \textbf{72.4} (88.1) & \textbf{97.5} & 90.2 (99.6)\\
 Geo + ShapeAug++  & 30 & 75.1 (96.7) & 68.3 (86.3) & \underline{97.4} & \underline{91.7} (100) \\
 Geo + ShapeAug++ & 50 & \underline{75.4} (96.6) & \underline{70.6} (87.7) & \underline{97.4} & \textbf{92.4} (100) \\
 
 \hline
\end{tabular}
\end{center}
\end{table*}

\subsection{Event Data Classification}

The comparison of ShapeAug++ to the original ShapeAug on multiple event classification datasets is presented in \autoref{tab:classification} for various maximum shape sizes $s_{max}$. ShapeAug++ outperforms ShapeAug on most datasets, with improvements of up to 3.7\% or at least maintaining equivalent performance. The difference in performance is particularly notable on more complex datasets, such as those involving real-world DVS recordings or a large number of classes. This demonstrates that the increased complexity of ShapeAug++ is especially beneficial for handling complex input data and provides a greater challenge for the prediction network. Similar to ShapeAug, we observe that the optimal maximum shape size still depends on the specific dataset, which is reasonable given that different datasets contain objects of varying sizes.

\subsection{Comparison on Robustness}

\begin{table*}[t]
\caption{Comparison of the robustness of current event augmentation approaches on various augmented versions of \textit{DVS Gesture} \cite{dvsgesture}. The best and second best results are shown in \textbf{bold} and \underline{underlined} respectively.}
\label{tab:robustness}
\begin{center}
\begin{tabular}{ccccccccc}
 \hline 
Train \ Valid  &- & Geo & Drop  &  Shape & \textbf{Shape++} \\
 \hline 
 Geo (Baseline) &  89.8 & 87.5 & 58.0 &  63.6 & 59.8\\
 \hline
 Geo + Drop & 89.8 & 87.9 & \textbf{86.0}& 73.9 & 50.8\\
 Geo + Mix & \textbf{93.2} & \textbf{92.4} & 82.6  &  76.9 & \underline{63.6}\\
 Geo + Shape & 91.7 & 90.5 & \underline{84.1} & \textbf{87.9} & 56.1 \\
Geo + \textbf{Shape++} & \underline{92.4} & \underline{92.0}  & 72.3 & \underline{84.8} & \textbf{85.6}\\
\hline
\end{tabular}
\end{center}
\end{table*}

We evaluate ShapeAug++ in comparison to other state-of-the-art methods using five challenging validation datasets based on the \textit{DVS Gesture} dataset \cite{dvsgesture}. These validation datasets were generated by applying the following augmentation techniques to each sample: Geometric transformations (horizontal flipping, rotation, cropping), EventDrop \cite{eventdrop}, ShapeAug \cite{shapeaug}, and ShapeAug++ with $s_{max}=30$. This allows us to assess if these augmentation techniques lead to an increased robustness of the trained neural networks. The results of our experiments are shown in \cref{tab:robustness}.

Our results indicate that ShapeAug++ significantly enhances robustness across all validation datasets compared to the baseline. It outperforms the original ShapeAug method on the original and geometrically transformed validation data. ShapeAug++ also demonstrates robustness against shape-augmented data. However, it shows a slight decrease in robustness against drop-augmentation, as its complex shapes differ from the masked-out squares used in EventDrop. The EventMix method achieves slightly higher performance than ShapeAug++ on some validation datasets. Nonetheless, these methods are not directly comparable, as EventMix introduces a multilabel classification problem.

Notably, all other augmentation methods show significantly lower robustness against validation data augmented with ShapeAug++, proving the necessity for more complex occlusion augmentation that closely resembles real-world scenarios.

\begin{table*}[t]
\caption{Evaluation of the robustness of the combination of different event augmentation methods on various augmented versions of \textit{DVS Gesture} \cite{dvsgesture}. The best and second best results are shown in \textbf{bold} and \underline{underlined} respectively.}
\label{tab:combination}
\begin{center}
\begin{tabular}{ccccccccc}
 \hline 
Train \ Valid  &- & Geo & Drop  &  Shape & \textbf{Shape++} \\
 \hline 
 Geo (Baseline) &  89.8 & 87.5 & 58.0 &  63.6 & 59.8\\
\hline
 Geo + Drop + Mix & 92.8 & 89.8 & \underline{89.8} & 75.4 & 64.0 \\
 Geo + Drop + Shape & 91.7 & 90.2 & 88.6 & 87.5 & 52.3 \\ 
 Geo + Mix + Shape & \underline{94.7} &  \underline{91.7} &  87.5 & \textbf{91.3} & 72.3\\
 Geo + Drop + Mix + Shape & \textbf{95.8} & \textbf{94.7} & \textbf{92.8} & 89.8 & 71.2\\
 Geo + Drop + \textbf{Shape++} & 90.2 & 86.7 & 86.4 & 83.3 & \underline{84.8}\\ 
 Geo + Mix + \textbf{Shape++} & 93.9 & 90.2 & 86.4 & 87.9 & \textbf{85.2}\\
 Geo + Drop + Mix + \textbf{Shape++} & 92.8 & 89.4 & 87.9 & 87.1 & \underline{84.8}\\
 \hline
\end{tabular}
\end{center}
\end{table*}

In \cref{tab:combination} we further explore the performance of combined augmentation strategies on the different validation datasets. The results demonstrate that incorporating EventMix into ShapeAug++ generally enhances performance across most data. Conversely, adding EventDrop reduces performance, even when combined with both EventMix and ShapeAug++. This might suggest that, due to the increased complexity of ShapeAug++, the network is too weak to handle further augmentation. Therefore, to effectively combine ShapeAug++ with other augmentation techniques, a different training schedule or a larger network may be necessary. Additionally, the results indicate that models trained without ShapeAug++ fail to perform well on ShapeAug++ augmented validation data, showing the importance of ShapeAug++ for handling complex real-world occlusions.

\section{Conclusion}

Augmentation techniques play a crucial role in enhancing the robustness and accuracy of neural networks, particularly for challenging tasks. ShapeAug introduced event data augmentation by simulating occlusions through the movement of square and circular objects along linear paths. In this work, we have extended this approach with ShapeAug++, introducing more complex, randomly shaped polygons and more realistic curved movements using Bézier curves. This improvement enables ShapeAug++ to more accurately model real-world scenarios, leading to greater robustness against occlusions.

Our experiments on the most commonly used event classification datasets demonstrate that ShapeAug++ achieves significantly higher accuracy than the baseline and outperforms the original ShapeAug method across most datasets. ShapeAug++ also exhibits strong performance on challenging validation datasets, surpassing not only ShapeAug but also the EventDrop method. Moreover, no current state-of-the-art augmentation technique is capable to achieve a high robustness against ShapeAug++ augmented validation data, showing the need for a more realistic occlusion simulation during the training. Despite its complexity, which may require more capable networks and extended training schedules, ShapeAug++ often delivers exceptional performance on its own, reducing the need for combining it with other augmentation techniques. 

ShapeAug++ currently applies occlusions randomly, without specifically targeting important objects in the data. Future research could explore integrating ShapeAug++ with saliency methods or other techniques to guide occlusions in a data-dependent manner, potentially enhancing its effectiveness even further.

\begin{credits}
\subsubsection{\ackname} 
This work was funded by the Carl Zeiss Stiftung, Germany under the Sustainable Embedded AI project (P2021-02-009).

\subsubsection{\discintname}
The authors have no competing interests to declare that are relevant to the content of this article.

\end{credits}

{\small
\bibliographystyle{splncs04}
\bibliography{egbib.bib}
}
\end{document}